\documentclass[runningheads]{llncs}
\usepackage{times}
\usepackage{soul}
\usepackage{url}
\usepackage[hidelinks]{hyperref}
\usepackage[small]{caption}
\usepackage{graphicx}
\usepackage{amsmath}
\usepackage{booktabs}
\usepackage{algorithm}
\usepackage{algorithmic}
\usepackage{multirow}
\usepackage{float}

\usepackage{tikz}
\usepackage{graphicx}
\usepackage{pgf-umlsd}  
\usetikzlibrary{shapes,backgrounds}
\usetikzlibrary{arrows,positioning}
\usetikzlibrary{trees}
\usetikzlibrary{chains}
\usepackage{schemabloc}
\usetikzlibrary{arrows,%
	petri,%
	topaths}%
\usepackage{tkz-berge}
\usepackage[position=top]{subfig}
\usepackage{mathpazo}
\usetikzlibrary{shadows}

\makeatletter
\newcommand{\printfnsymbol}[1]{%
	\textsuperscript{\@fnsymbol{#1}}%
}
\makeatother
\begin{document}
\newcommand{\mvspace}{\setlength\intextsep{0pt}} 
\authorrunning{Sun, Lin, and Bischl}
\titlerunning{ML ReinBo}
\title{ReinBo:  Machine Learning pipeline search and configuration with Bayesian Optimization embedded Reinforcement Learning}
\author{Xudong Sun \thanks{equal contribution.}  \and
	Jiali Lin\printfnsymbol{1} \and
	Bernd Bischl}
%
%
\institute{LMU Munich}
\maketitle              
\begin{abstract}
Machine learning pipeline potentially consists of several stages of operations like data  preprocessing, feature engineering and machine learning model training. Each operation has a set of hyper-parameters, which can become irrelevant for the pipeline when the operation is not selected. This gives rise to a hierarchical conditional hyper-parameter space. To optimize this mixed continuous and discrete conditional hierarchical hyper-parameter space, we propose an efficient pipeline search and configuration algorithm which combines the power of Reinforcement Learning and Bayesian Optimization. Empirical results show that our method performs favorably compared to state of the art methods like Auto-sklearn , TPOT, Tree Parzen Window, and Random Search.  
\end{abstract}

\section{Introduction}
Over the past years, Machine Learning (ML) has achieved remarkable success in a wide range of application areas, which has greatly increased the demand for machine learning systems. However, an efficient machine learning algorithm crucially depends on a human expert, who has to carefully design the pipeline of the machine learning system, potentially consisting of data pre-process, feature filtering, machine learning algorithm selection, as well as identifying a good set of hyper-parameters. As there are a large number of possible alternatives of models as well as hyper-parameters, the need for automated machine learning (AutoML) has been growing, which is supposed to automatically generate a data analysis pipeline with machine learning methods and parameter settings that are optimized for a given data set, in order to make machine learning methods available for non-expert users.

Since hyper-parameters of a machine learning model have a large influence on the performance of the model, hyper-parameter optimization becomes a critical part of an AutoML system. Popular hyper-parameter optimization methods include Sequential Bayesian Optimization,  which iterates between fitting surrogate models that predict model performance, and using them to make choices about which configurations to investigate. 

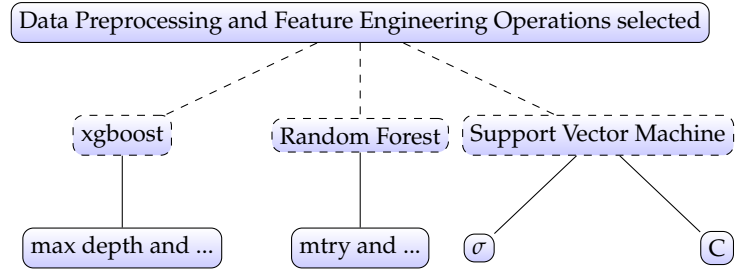
\begin{figure}
	\begin{center}
\begin{tikzpicture}[sibling distance=10em,
every node/.style = {shape=rectangle, rounded corners,
	draw, align=center,
	top color=white, bottom color=blue!20}]]

\node {Data Preprocessing and Feature Engineering Operations selected}
child[dashed]{ node {xgboost} 
	child[solid]{node {max depth and ...}}
}
child[dashed]{ node {Random Forest} 
	child[solid]{node {mtry and ...}}
}
child[dashed]{ node {Support Vector Machine}
	child[solid]{ node {$\sigma$}}
	child[solid]{ node {C} } };
\end{tikzpicture}
\caption{Example of Conditional Hierarchical Space}
\label{fig:cash}
		
	\end{center}
\end{figure}

However, the composition of the machine learning pipelines also plays a vital role in the performance of AutoML systems. Choosing different data preprocessing or feature engineering techniques as well as choosing different machine learning models for a specific dataset could potentially result in considerable performance differences. The joint optimization of the pipeline search and its associated hyper-parameters configuration could essentially reside under the umbrella of Combined Algorithm Selection and Hyperparameter optimization (CASH) problem \cite{thornton2012auto}, where Algorithm corresponds to the pipeline and Configuration corresponds to the hyper-parameters associated with the pipeline. The pipelines and hyper-parameters reside in a conditional hierarchical space, where some hyper-parameters only become valid when the corresponding pipeline is present. For example, Figure \ref{fig:cash} illustrates such a situation when the data preprocessing and feature engineering operations are selected, which correspond to an incomplete pipeline, one out of three machine learning algorithms need to be chosen (indicated by dashed edges) to complete the pipeline, the corresponding hyper-parameters (indicated by solid edges) of an algorithm only become valid when the algorithm is selected.

To optimize the conditional hyper-parameters space jointly with the pipeline it is attached to, we embed Bayesian Optimization in the Reinforcement Learning process, and dub the method ReinBo, which means Machine Learning Pipeline search and configuration with Reinforcement Learning and Bayesian Optimization. Note that ReinBo can solve not only CASH problems, but also any mixed discrete and continuous conditional hierarchical space optimization, which is left for future work. 

Our major contributions are:
\begin{itemize}
\item Inspired by Hierarchical Reinforcement Learning \cite{dietterich2000hierarchical}, we transform the conditional hierarchical hyper-parameter optimization problem into subtasks of pipeline selection and hyper-parameter optimization, which circumvents the conditional constraint and reduces the search dimension.
\item To our best knowledge, we are the first to embed Bayesian Optimization (BO) into Reinforcement learning, specifically Q Learning \cite{watkins1992q} for collaborative joint search of pipelines and hyper-parameters, which is different from using BO for policy optimization \cite{brochu2010tutorial}, and also different from using BO for hyper-parameter fine tuning after an optimal pipeline is selected by a reinforcement learning based AutoML framework \cite{yang2019program}. 
\item We provide an open source light weight R language implementation \textit{reinbo}\footnotemark\footnotetext{https://github.com/smilesun/reinbo} for the R Machine Learning community which could run efficiently on a personal computer, and takes much less resources compared to other AutoML softwares. \end{itemize}
 
In the following section, we review related works and discuss the differences to our method. In section \ref{sec:method}, we explain our method in detail and also shed light to connections with Hyperband \cite{li2017hyperband}. In section \ref{sec:exp}, we benchmark our method by comparing it with several state of art methods.

\section{Related Work}
In this section, we try to classify the current popular AutoML solutions into a taxonomy and discuss the differences of each individual work with ours.

\textbf{Sequential Model Based Optimization family}\,  Auto-sklearn \cite{feurer2015efficient} and Auto-Weka \cite{thornton2012auto} both use Sequential Model-based Algorithm Configuration (SMAC) \cite{hutter2011sequential} to solve the Combined Algorithm Selection and Hyperparameter optimization (CASH) problem. SMAC\cite{hutter2011sequential} transforms the conditional hierarchical hyper-parameter space into a flat structure by instantiating inactive conditional parameters to default values, which allows the random forest to focus on active hyper-parameters \cite{hutter2011sequential}. A potential drawback for this method is that the surrogate model needs to learn in a high dimensional hyper-parameter space, which might need a large sample of observations to be sufficiently trained, while in practice, running machine learning algorithm is usually very expensive. Tree Parzen Window (TPE) \cite{bergstra2011algorithms}, however, tackles the conditional hierarchical hyper-parameter space using a tree like Parzen Window to construct two density estimators on top of a tree like hyper-parameter set. Expected improvement induced from lower and upper quantile density estimators is used to select new candidate proposals from points generated by Ancestral Sampling.

\textbf{Tree-based Genetic Programming}\, TPOT \cite{olson2016tpot} automatically designs and optimizes machine learning pipelines with a genetic programming \cite{banzhaf1998genetic} algorithm. The machine learning operators are used as genetic programming primitives, which will be combined by tree-based pipelines and the Genetic Programming algorithm is used to evolve tree-based pipelines until the “best” pipeline is found. Similar methods also include Recipe \cite{de2017recipe}. However, this family of methods does not scale well \cite{mohr2018ml}. In this paper, we aim for an AutoML system that could give a valuable configured pipeline within limited time.

\textbf{Monte Carlo Tree Search Alike}\, ML-Plan \cite{mohr2018ml} is an AutoML system, built upon a Hierarchical Task Network, which uses a Monte Carlo Tree Search like algorithm to search for pipelines and also configure the pipeline with hyper-parameters. Task is expanded based on best-first search, where the score is estimated by a randomized depth first search by randomly trying different subtree possibilities on a Hierarchical Task Network. 
To ensure exploration, ML-Plan gives equal possibility to the starting node in a Hierarchical Task Network and then uses a best-first strategy for searching at the lower part of the network. 
Potential drawback for this method is that the hyper-parameter space is discretized, which might essentially lose good candidates in continuous spaces since large continuous hyper-parameter spaces would be essentially hard to discretize. 

\textbf{Reinforcement Learning based Neural Network Architecture Search}\, This family of methods are usually not termed as AutoML systems but rather Neural Architecture Search. For instance, MetaQNN \cite{baker2017designing} uses Q-learning to search convolutional neural network architectures. The learning agent is trained to sequentially choose CNN layers using Q-learning with an $\epsilon$-greedy exploration strategy and the goal is to maximize the cross-validation accuracy. 
In \cite{zoph2016neural}, instead of using Q-learning, the authors use Recurrent Neural Networks as the reinforcement learning policy approximator to generate variable strings to represent various neural architecture forms. The reward-function is designed to be the validation performance of the constructed network. The reinforcement learning policy is trained with gradient descent algorithm, specifically REINFORCE. The architecture elements being searched are very similar to MetaQNN. Inspired from \cite{zoph2016neural}, we also assume the machine learning pipeline to be optimized could be represented by a variable length string, but our work is different from \cite{zoph2016neural} in that we use both Deep Q-learning and Tabular Q-learning. More importantly, compared with both \cite{baker2017designing} and \cite{zoph2016neural}, which optimize the neural architecture, the elements of the architecture are mostly factor variables like layer type or discretized elements like filter size, while in this paper, we deal heavily with continuous hyper-parameters (The $C$ and $\sigma$ for a rbf kernel Support Vector Machine). To jointly optimize the discrete pipeline choice and associated continuous hyper-parameters, we embed Bayesian Optimization inside our reinforcement learning agent. 

\textbf{Other Reinforcement Learning based methods}\, In
\cite{yang2019program}, the authors also combine pipeline search and  hyper-parameter optimization in a reinforcement learning process based on the PEORL \cite{yang2018peorl} framework, however, the hyper-parameter is randomly sampled during the reinforcement learning process, an extra stage is needed to sweep the hyper-parameters using hyper-parameter optimization techniques, while in our work, hyper-parameter optimization is embedded in the reinforcement learning process. 
Alpha3M \cite{drori2018alphad3m} combined MCTS and recurrent neural network in a self play \cite{silver2017mastering} fashion, however, it seems that Alpha3M does not perform better than the state of art AutoML systems. For example, out of all the 6 OpenML datasets they have used to compare with state of art AutoML systems, Alpha3M only shows a clear improvement on 1 dataset (spectf) against Auto-sklearn \cite{feurer2015efficient} and TPOT \cite{olson2016tpot}, according to Figure 4 in \cite{drori2018alphad3m}. Furthermore, 
it is not clear to us how the hyper-parameters are set and if Bayesian Optimization is used. The implementation of Alpha3M takes advantage of the GPUs \cite{drori2018alphad3m} for the fast performance while our method has a light weight implementation which efficiently runs with CPU and does not necessarily need Neural Networks.   
\section{Method}
\label{sec:method}
\subsection{Towards Reinbo}
\label{subsec:reinbo}

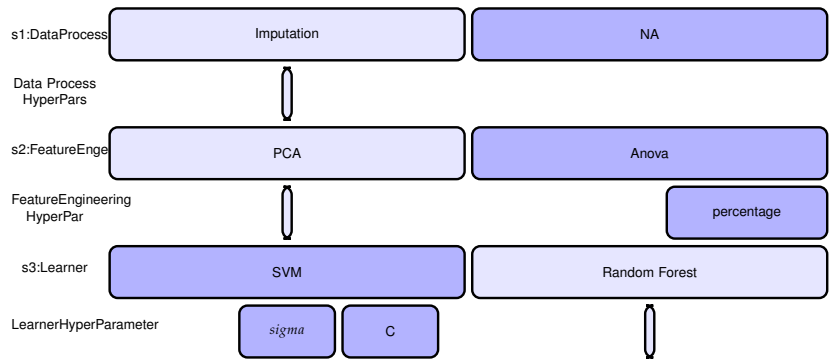
\begin{figure}
	\begin{center}
	\resizebox{.9\linewidth}{!}{
\begin{tikzpicture}[
	scale=0.65,
	start chain=1 going below, 
	start chain=2 going right,
	node distance=1mm,
	desc/.style={
		scale=0.75,
		on chain=2,
		rectangle,
		rounded corners,
		draw=black, 
		very thick,
		text centered,
		text width=8cm,
		minimum height=12mm,
		fill=blue!30
		},
	it/.style={
		fill=blue!10
	},
	level/.style={
		scale=0.75,
		on chain=1,
		minimum height=12mm,
		text width=2cm,
		text centered
	},
	every node/.style={font=\sffamily}
]


\node [level] (Level 3) {s1:DataProcess};
\node[level](dphp) {Data Process HyperPars};
\node [level] (Level 2) {s2:FeatureEngeering};
\node [level] (Level 1.5) {FeatureEngineering HyperPar};
\node [level] (Level 1) {s3:Learner};
\node [level] (Level 0) {LearnerHyperParameter};

\chainin (Level 3); 
\node[desc, it](impute){Imputation};
\node[desc](dpna){NA};
\chainin (impute);
\node [desc, it, text width=0cm, continue chain=going below] (fehp1) {};
\chainin (fehp1);
\node[desc,it, continue chain=going below](pca){PCA};
\node [desc, continue chain = going right] (anova) {Anova};
\chainin (pca);
\node [desc, it, text width=0cm, continue chain=going below] (fehp2) {};
\node [desc] (svm) {SVM};
\node [desc, it, continue chain=going right] (rf) {Random Forest};
\node [desc, it, text width=0cm, continue chain=going below] (fehp2) {};
\chainin(svm);
\node [desc, text width=2cm, xshift=0, continue chain = going below] (sigma) {$sigma$};
\node [desc, text width=2cm, xshift=0, continue chain = going right] (c) {C};
\chainin (anova);
\node [desc, text width=3.5cm, xshift=2.25cm, continue chain = going below] (PLC) {percentage};
\end{tikzpicture}}
\end{center}
\caption{Toy example of selected pipeline and associated hyperparameters}
\label{fig:selpipe}
\end{figure}
As shown in Figure \ref{fig:selpipe}, we assume that a machine learning pipeline potentially consists of 3 stages (s1 through s3 in the figure), which include data preprocessing (imputations, NA and more), feature engineering (Principal Component Analysis for feature transform, Anova for feature filtering and more), and machine learning model (learner like SVM, Random Forest). Specifically, we use operation ``NA'' to indicate that no operation would be done in the stage in question. Figure \ref{fig:selpipe} just serves as a toy but working example for ReinBo, in practice, there are a lot more operations available. A particular operation has associated hyper-parameters (for instance the percentage of selected features in Anova feature filtering). In Figure \ref{fig:selpipe}, dark color filled cells (NA, Anova, SVM) represent selected operations and their associated active hyper-parameters (percentage, sigma, C), while hyper-parameters for inactive operations are not drawn in the figure. 

Observing from Figure \ref{fig:selpipe}, along with Figure \ref{fig:cash}, we could think of the pipeline selection and configuration problem as a two-phase process. During the first phase, a planning algorithm guides the agent to choose a path which corresponds to an unconfigured pipeline. This is similar to a multi-armed bandit problem, where each path corresponds to one arm, while difference lies in that the agent can not directly pull a discrete arm but have to pull across several consecutive discrete arm groups (each arm group corresponds to a stage in Figure \ref{fig:selpipe}) and the agent only gets reward after choosing one of arms from the last group. The second phase is similar to contextual bandit with continuous action space (corresponding to hyper-parameters), where the context is which path from the first phase has been selected.

We model the first phase as a reinforcement learning episode, where a particular operation in stage $i$ is treated as action $a_i$, taken upon corresponding state $s_i$. State $s_i$ could be represented by actions taken up to the current stage for example. The pipeline search problem is then to find an optimal policy $\pi$ to decide which operation (action) to take at a particular state. The action value function $Q(s, a)$ at each state tells us how favorable a particular operation is. We use $\mathcal{A}_{s_i}$ to denote the space of legal actions at state $s_i$. Suppose a roll-out of states trajectory for one composition (episode) is $s_1,\hdots, s_K$, the corresponding space of pipeline could be denoted by $\prod_{i=1}^K \mathcal{A}_{s_i}$, where $K$ is the total number of stages and we use $\prod$ to denote the Cartesian Product. For a more general notation, we use $\mathcal{A}(S_i)$ to denote the space of actions when the state is $S_i$ at stage $i$.

We search for potentially better hyperparameters in the second phase with Bayesian Optimization.
Aside from the pipeline itself, each concrete operation (action $a_i$) at stage $i$ is configurable by a set of hyper-parameters $\Phi_{a_i}$. $\Phi_{a_i}$ can be hyper-parameters set for a preprocessor like the ratio of variance to keep in PCA or hyper-parameters set for a machine learning model like the $C$ and $\sigma$ hyper-parameter for SVM. Thus a configured pipeline search space would be $\prod_{i=1}^K \mathcal{A}(S_i;\Phi_{a_i})$ where we use $\Phi_{a_i}$ to denote the conditional hyper-parameter space at stage $i$. 


The connection point between reinforcement learning and Bayesian Optimization lies in the reward function design in the reinforcement learning part. During the composition process, there is no signal available to judge how good a current uncompleted pipeline is until the final learner (classifier) is configured with hyper-parameters and trained on the data. At the starting point, different pipelines are tried out randomly, which corresponds to an untrained exploration policy $\pi$. A completed pipeline with a joint non-conditional hyper-parameter search space is optimized with Bayesian Optimization for a few steps. 
The best negative loss is then used as a reward at the end of an episode to guide the reinforcement learning agent towards a better policy. The environment uncertainty only comes with the stochastic reward, while the transition from current state to next state through action is deterministic.
We choose to use Q-learning \cite{watkins1992q} to optimize the policy where we have tried the Tabular Q-learning and Deep Q-learning \cite{mnih2015human}. We find out that the Tabular Q-learning works better than Deep Q-learning. For space constraint, the latter is not discussed in detail in this work.



We need Bayesian Optimization to optimize the hyper-parameters in a fine grained level with limited budget, but also want to give budget preference to those promising pipelines. To circumvent the complexity of conditional and hiercharical relationship between hyper-parameters and pipeline, we use reinforcement learning to choose a pipeline and let Bayesian Optimization tune the hyper-parameters. We model the variation of the same pipeline with different hyper-parameters as the environment uncertainty. By using separate surrogate model for each pipeline, we circumvent the risk of mistakenly modeling improper dependent structure between different hyper-parameters, at a minor cost of maintaining those searched pipelines surrogate model as dictionary in memory. 

\subsection{Connections to Hyperband}
The idea of only using a few steps Bayesian Optimization is inspired by Hyperband \cite{li2017hyperband}, where the trade-off between aggressively exploring more configurations and giving each configuration more resources to be validated is solved by grid searching. Instead, in this paper, we do not need the grid search, promising pipelines will get a higher probability to be selected by our reinforcement learning agent which means these pipelines get more chances to be evaluated by the Bayesian Optimization process. The trade-off between exploitation and exploration is naturally resolved by an $\epsilon$-greedy policy, and by annealing $\epsilon$ from a large value to a small value, we encourage more exploration at the beginning. Compared to Hyperband, our method selects the budgets allocated for a particular pipeline automatically, the effectiveness of our strategy could then rely on the recent success of reinforcement learning in different areas.
\subsection{Connection and Extension to Hierarchical Reinforcement Learning}
Hierarchical Reinforcement Learning (hrl) \cite{barto2003recent} is proposed to tackle the curse of dimensionality in Reinforcement Learning \cite{kulkarni2016hierarchical}. Although the Option  approach \cite{barto2003recent} is more popular, our method has a close connection to the MAXQ subtask approach \cite{dietterich2000hierarchical}, which divide a task recursively into subtasks and decompose the value function accordingly. The current version of ReinBo can be treated as a special case of the MAXQ task decomposition, where we have two tasks of pipeline selection and hyper-parameter configuration. However, in the current version, most states are not shared between these two tasks, so there is no need to use MAXQ hrl algorithm to solve the problem. But our method can be naturally extended to a hrl version when our design space of pipeline allow shared state between the two subtasks. We leave it as future work to optimize such complicated pipelines using Hierarchical Reinforcement Learning.
\subsection{Procedures of ReinBo}
As shown in Algorithm \ref{algo:reinbo} , we first initialize a policy $\pi$ for the agent which can be represented by a randomly initialized neural network or a Q-table initialized with default values, coupled with an exploration mechanism like the $\epsilon$-greedy strategy. During the roll-out, initial populations of pipelines get sampled, with the corresponding hyper-parameter space $\Lambda (\prod_i \Phi_{a_i})$ to be optimized by Bayesian Optimization for several steps. The corresponding surrogate model is stored in the dictionary $\mathcal{R}$ for future episode if the same pipeline gets rolled out again. The performance of the pipeline on validation data will be used to serve as feedback signal or reward to the reinforcement learning agent to conduct policy iteration. 
\begin{algorithm}
\caption{ML Reinbo}
\begin{algorithmic}
\REQUIRE dataset $\mathcal{D}$, Operators and Hyper-parameters\\
Initialize Policy $\pi$\\
Initialize MBO Surrogate Dictionary $\mathcal{R} \leftarrow \emptyset$ \\
\WHILE{Budget not reached}
\STATE Roll-out an \textbf{unconfigured} pipeline $\prod a_i$ according to policy $\pi$\\
\STATE Get hyper-parameters set for the ground pipeline $\Lambda(\prod a_i) = \prod_i \Phi_{a_i}$ \\
\STATE Reward $R \leftarrow  \text{MBO\_PROBE}(\prod a_i, \Lambda, \mathcal{R})$ 
\STATE Update Policy $\pi$ with reinforcement learning algorithm
\ENDWHILE 
\end{algorithmic}
\label{algo:reinbo}
\end{algorithm}

Once an unconfigured pipeline is constructed at the end of the episode, running Bayesian Optimization could be beneficial in searching for a more favorable hyper-parameter setting. However, Bayesian Hyperparameter Optimization with large budgets could be rather expensive. Instead, we optimize hyper-parameters for an un-configured pipeline only for several iterations. For example, we take the number of iterations to be 2 or 3 times the dimension of hyper-parameter space, which means that hyper-parameter spaces with higher dimension will get more sampling budgets. After each episode, the current best configuration's performance for this pipeline in question is used as reward. The next time the same pipeline is sampled, the surrogate model could be retrieved from the dictionary $\mathcal{R}$ to facilitate further search using MBO. We dub the hyper-parameter search process as MBO\_PROBE, with details shown in Algorithm \ref{algo:mbo-prob}.\footnotemark \footnotetext{
To save budgets, when an un-configured pipeline does not improve after a number of trials of MBO\_PROBE, it can also be suspended for future evaluation.}
If an un-configured pipeline is not sampled yet, an initial design is generated to facilitate an initial surrogate model.

\begin{algorithm}
	\caption{MBO\_PROBE($\prod a_i, \Lambda, \mathcal{R}$)}
	\begin{algorithmic}
		\REQUIRE ~~\\
		\IF{$\mathcal{R}\{\prod_i a_i\} = \emptyset$} 
		\STATE generate initial design of size $n^{init}$ for surrogate model; \\
		\FOR{$j$ in $1:n^{init}$}
			\STATE evaluate each design by initiating the pipeline with corresponding hyper-parameters $\Lambda(\prod a_i)$.\\
		\ENDFOR
		\STATE initialize $\mathcal{R}\{\prod_i a_i\}$
	    \ENDIF
		\FOR{$j$ in $1:n^{probe}$} 
		\STATE propose new point according to surrogate model $\mathcal{R}\{\prod a_i\}$
		\STATE evaluate new point
		\ENDFOR
		
		\RETURN $y$  $\leftarrow$ best accuracy until now
	\end{algorithmic}
\label{algo:mbo-prob}
\end{algorithm}
\section{Experiments}
\label{sec:exp}

\subsection{Implementation, Comparision Methods and Setups}
\label{subsec:baseline}
Our initial implementation for ReinBo is based on R machine learning packages \textit{mlr}~\cite{mlr}, \textit{mlrCPO} \cite{cpo} for pipeline construction and \textit{mlrMBO} \cite{bischl2017mlrmbo} for Bayesian Optimization. However, ReinBo could be extended to combine with any machine learning library as backends. The R package \textit{parabox}\footnotemark{}\footnotetext{https://github.com/smilesun/parabox} is implemented for this project to specify conditional hierarchical hyper-parameter space and provides the conditional ancestral sampling (random search in conditional hyper-parameter space). The R package \textit{rlR}\footnotemark{}\footnotetext{https://github.com/smilesun/rlR} is implemented for reinforcement learning where the user could implement a custom environment as input. All python packages are invoked with the R-Python interface \textit{reticulate} \cite{reticulate}.

To evaluate the performance of our proposed method, we compare the performance of ReinBo with several state of art AutoML systems, as well as several conditional hyper-parameter space tuning methods. We compare against Auto-sklearn \cite{feurer2015efficient} and TPOT~\cite{olson2016tpot}, both based on \textit{scikit-learn}~\cite{scikit-learn}.
At the time of writing, there is no documentation and examples online for ML-Plan \cite{mohr2018ml}, thus, ML-Plan is not included in the experiment. Additionally, we compare against hyperparameter optimization techniques which preserve the hierarchical conditional structure, including Tree-structured Parzen Estimator (TPE)~\cite{bergstra2011algorithms} used in \textit{Hyperopt}~\cite{bergstra2013hyperopt}, and Random Search with conditional Ancestral Sampling (self implemented in R package \textit{parabox}). Random Search remains a very strong baseline in a lot of machine learning hyper-parameter optimization scenarios \cite{bergstra2012random}. 

\textbf{Evaluation Criteria}\,
As warned in \cite{mohr2018ml}, many state of art AutoML systems seem to have missed to deal with the risk of overfitting. Therefore, in the experiment part, we focus on evaluating the generalization capability of the selected pipeline empirically. To avoid any potential confusion from synonyms, we use $D^{opt}$ to represent the part of a dataset fed into a given AutoML system and use $D^{test}$ to represent the locked out part \cite{sun} of the same dataset used to test the generalization capacity. The split of $D^{opt}$ and $D^{test}$ is done by Cross Validation, which means for a dataset $D$, $D = D^{opt} \bigcup D^{test}$ and $D^{opt} \bigcap D^{test} = \emptyset$. To create the $D^{opt}$ and $D^{test}$ split, we use 
5-fold cross-validation (\textit{CV5}), which corresponds to the outer loop of Nested Cross Validation (\textit{NCV}). We take the aggregated mmce (mean miss-calssification error) across the 5 iterations of outer loop of \textit{NCV} as performance measure.


Instead of using running time as budget, we use the number of pipeline configurations as the unit of budget, to account for performances variations due to hardware and network load etc. For $D^{opt}$, we assigned a budget of 1000 times of \textit{CV5} equivalent to each AutoML algorithm, which corresponds to the inner loop of \textit{NCV}. 

\textbf{Other Setups}\,
To account for different AutoML systems data input format incompatibility problem, we conduct dummy encoding to categorical features beforehand. Aiming for a light weight implementation, in the experiment, we limit our choice of pipeline components for ReinBo. We compose a pipeline in 3 stages, with potential operations/actions at each stage listed in Table \ref{tb:expipe}. Associated hyper-parameters with an un-configured pipeline are listed in Table \ref{tb:hyperspace}. We call the components and associated hyper-parameters the pipeline pool. The same pipeline pool is used for ReinBo, TPE and Random Search. 

For Auto-sklearn, Meta-learning and ensemble are disabled, the resampling strategy is set to be \textit{CV5}, stop criteria is changed to budget instead of time and all other configurations are kept default. We have contacted the author of Autosklearn through github for the right use of the API to ensure the above configuration is satisfied. 
For TPOT (version 0.9), the default configuration space contains a lot of operators while the light version provides only fast models and pre-processors. The light TPOT is therefore less time-consuming but it could probably lead to lower accuracy in consequence. For this reason, we compare ReinBo with both TPOT with the default configuration and TPOT with light configuration, and we call them TPOT and TPOT-light respectively. TPOT is configured to allow equal amount of budgets with all methods being compared and other configurations are left to be default.



%


\textbf{Datasets}\,
We experimented on a set of standard benchmarking datasets of high quality collected from OpenML-CC18\footnotemark{}\footnotetext{https://www.openml.org/s/99}~\cite{bischl2017openml} and Auto-Weka ~\cite{DBLP:journals/corr/abs-1208-3719}, which are rather well-curated from many thousands and have diverse numbers of classes, features, observations, as well as various ratios of the minority and majority class size. Summary of these datasets is listed in Table \ref{tb:data}.
\begin{table}[p]\small 
	\caption{List of pipeline operations. An operation of ``NA'' here is used to indicate that no operation would be taken in the corresponding stage. Please refer to \textit{mlrCPO} documentation for the detailed meaning of these operators.}
	\centering
	\footnotesize
	\scalebox{0.75}{
		\begin{tabular}{cl|ccccc}	
			\toprule
			& \textbf{Stage}& \multicolumn{5}{c}{\textbf{Operation/Action}} \\
			\midrule
			1& DataPreprocess&  	Scale(default)& Scale(center=FALSE)& Scale(scale=FALSE)&  SpatialSign&         NA\\
			2& Feature Engineering&      Pca&             FilterKruskal&	             FilterAnova&	  FilterUnivariate&	      NA\\
			3& Classifier&	     kknn&                ksvm&          	ranger&	         xgboost& naiveBayes\\
			\bottomrule
	\end{tabular}}
	\label{tb:expipe}
\end{table}
\begin{table}[p]
	\caption{List of hyper-parameters to the operations in Table \ref{tb:expipe}. ``p'' in the column ``Range'' indicates the number of features of the original dataset. We invite the user to refer to the R packages \textit{mlrCPO} and \textit{mlr} documentations for the exact meaning of operation, hyper-parameters, etc.}
	\centering
	\footnotesize
		\scalebox{0.75}{
	\begin{tabular}{lllr}
		\toprule
		\textbf{Operation}&\textbf{Parameter}&\textbf{Type}&\textbf{Range}\\
		\midrule
		Anonva,Kruskal,Univariate&               perc&	    numeric&              (0.1,1)\\
		                      Pca&               rank&      integer&             (p/10,p)\\
		                     kknn&                  k&      integer&               (1,20)\\
		                     ksvm&                  C&	    numeric& ($2^{-15}$,$2^{15}$)\\
		                     ksvm&              sigma&	    numeric& ($2^{-15}$,$2^{15}$)\\
		                   ranger&               mtry&	    integer&         (p/10,p/1.5)\\
		                   ranger&    sample.fraction&	    numeric&              (0.1,1)\\
		                  xgboost&                eta&	    numeric&          (0.001,0.3)\\
		                  xgboost&         max\_depth&	    integer&               (1,15)\\
		                  xgboost&          subsample&	    numeric&              (0.5,1)\\
		                  xgboost&   colsample\_bytree&      numeric&              (0.5,1)\\
		                  xgboost&  min\_child\_weight&      numeric&               (0,50)\\
		               naiveBayes&             laplace&      numeric&         (0.01,100)\\
		\bottomrule
	\end{tabular}}
\label{tb:hyperspace}
\end{table}
\begin{table}
	\centering
	\caption{List of OpenML Datasets for experiment. Columns are the OpenML task\_id and name, the number of classes (nClass), features (nFeat) and observations (nObs), as well as the ratio of the minority and majority class sizes (rMinMaj).}
	\footnotesize
	    \scalebox{0.75}{
	    \begin{tabular}{cccccc}
	    \toprule
	    \textbf{task\_id} & \textbf{name}& \textbf{nClass} & \textbf{nFeat} & \textbf{nObs} & \textbf{rMinMaj}\\
	    \midrule
	    14& mfeat-fourier& 10& 77& 2000& 1.00\\
	    23& cmc& 3& 10& 1473& 0.53\\
	    37& diabetes&	2&	9&	768&	0.54\\
	    53&	vehicle& 4&	19&	846&	0.91\\
	    3917& kc1& 2& 22& 2109& 0.18\\
	    9946&	wdbc& 2&	31&	569&	0.59\\
	    9952& phoneme& 2& 6& 5404& 0.42\\
	    9978& ozone-level-8hr& 2& 73& 2534& 0.07\\
	    125921&	LED-display-domain-7digit& 10&	8&	500&	0.65\\
	    146817& steel-plates-fault&	7&	28&	1941& 0.08\\
	    146820& wilt& 2& 6& 4839& 0.06\\
	    \bottomrule
	    \end{tabular}
	    }
    \label{tb:data}
\end{table}

\subsection{Experiment results}
\begin{figure}[p]
\centering
\includegraphics[scale=0.6]{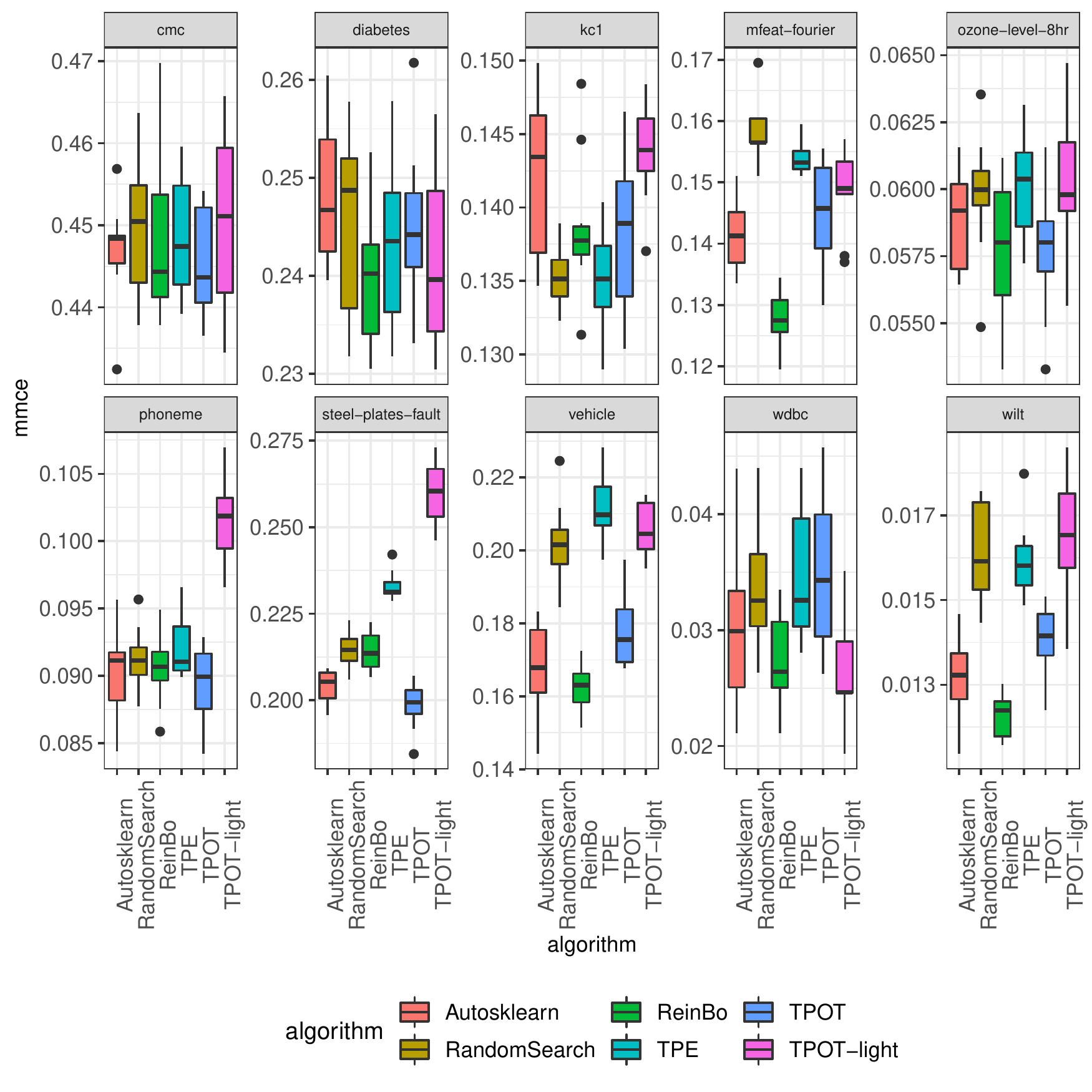}
\caption{Boxplots showing the distribution of aggregated mmce achieved by each algorithm within 10 statistical replications.} 
\label{fig:lock}
\end{figure}

In Figure \ref{fig:lock}, we compare the mmce (1-Accuracy) of each method with boxplot over the datasets listed in Table \ref{tb:data} across 10 statistical replications. Additionally, we list numerical results in Table \ref{tab:ncv} with statistical test, where each numerical value represents the aggregated mean mmce over the statistical replications. Underline in each row indicates the smallest mean value over the corresponding dataset. The bold-faced values indicate that the corresponding algorithm does not perform significantly worse than the underlined algorithm on the corresponding dataset based on Mann-Whitney U test. As shown in Table \ref{tab:ncv}, ML-ReinBo has boldfaces for 8 of 10 datasets followed by much less boldfaces from other methods.

In Table \ref{tab:test}, we compare win (significantly better), lose and tie (neither significantly better nor worse) relationships according to the test.
As shown in Table \ref{tab:test}, ReinBo has won TPOT on 5 datasets and performed worse than TPOT for only one dataset. And not surprisingly, TPOT has performed considerably better than TPOT-light in the empirical experiments since TPOT-light has smaller search space with only fast models and preprocessors. ReinBo also performs much better than Random Search and TPE, where ReinBo has significantly won them on 5 and 6 tasks respectively and lost only on 1 task. Compared to ReinBo, Auto-sklearn has won only once and behaved worse than ReinBo on 3 of 10 datasets. 

\begin{table}
	\centering
\caption{Average performance (mmce) of algorithms across 10 statistical replications with different seeds. In each run the aggregated mmce based over the outer loop of \textit{NCV} is taken as performance measure for each algorithm. Each value in this table is the mean value of the aggregated mmce values across 10 replications and the best mean value for each dataset is underlined. The bold-faced values indicate that the algorithm does not perform significantly worse than the underlined algorithm on the corresponding dataset based on Mann-Whitney U test.}	
\footnotesize
\scalebox{0.75}{
\begin{tabular}{c|c|c|c|c|c|c|c}
	\hline
	dataset & Auto-sklearn & TPE & TPOT & TPOT-light & Random & ReinBo & Underlined Algorithm\\
	\hline
	mfeat-fourier & 0.1412 & 0.1542 & 0.1451 & 0.1489 & 0.1580 & \underline{\textbf{0.1278}} & ReinBo\\
	cmc & \textbf{0.4470} & \textbf{0.4485} & \underline{\textbf{0.4457}} & \textbf{0.4506} & \textbf{0.4500} & \textbf{0.4485} & TPOT\\
	diabetes & 0.2483 & \textbf{0.2436} & 0.2452 & \textbf{0.2413} & \textbf{0.2455} & \underline{\textbf{0.2395}} & ReinBo\\
	vehicle & \textbf{0.1679} & 0.2117 & 0.1784 & 0.2057 & 0.2020 & \underline{\textbf{0.1621}} & ReinBo\\
	kc1 & 0.1421 & \underline{\textbf{0.1351}} & \textbf{0.1380} & 0.1438 & \textbf{0.1353} & 0.1387 & TPE\\
	wdbc & \textbf{0.0299} & 0.0348 & 0.0353 & \underline{\textbf{0.0264}} & 0.0341 & \textbf{0.0271} & TPOT-light\\
	phoneme & \textbf{0.0902} & \textbf{0.0920} & \underline{\textbf{0.0893}} & 0.1016 & \textbf{0.0912} & \textbf{0.0905} & TPOT\\
	ozone-level-8hr & \textbf{0.0588} & 0.0601 & \underline{\textbf{0.0577}} & 0.0603 & 0.0598 & \textbf{0.0578} & TPOT\\
	steel-plates-fault & 0.2041 & 0.2330 & \underline{\textbf{0.1985}} & 0.2601 & 0.2146 & 0.2141 & TPOT\\
	wilt & 0.0132 & 0.0159 & 0.0141 & 0.0164 & 0.0161 & \underline{\textbf{0.0123}} & ReinBo\\
	\hline
\end{tabular}
}
\label{tab:ncv}
\end{table}

\begin{table}
	\centering
	\caption{Win-Lose-Tie comparison between ReinBo and other algorithms on benchmarking datasets based on Mann-Whitney U test (significance level $\alpha = 0.05$).}
	\scalebox{0.75}{
	\begin{tabular}{cc|ccccc}
	    \toprule
	    & & \textbf{Random\_search} & \textbf{TPE} & \textbf{Auto-sklearn} & \textbf{TPOT-light} & \textbf{TPOT}\\
	    \midrule
	    \multirow{3}{*}{\textbf{ReinBo}} 
	    & win& 5& 6& 3& 7& 5\\
	    & tie& 4& 3& 6& 3& 4\\
	    & lose& 1& 1& 1& 0& 1\\
	    \bottomrule
	\end{tabular}}
	\label{tab:test}
\end{table}
Meanwhile, ReinBo has comparatively short box ranges in most cases as shown in Figure \ref{fig:lock}. Hence, we would conclude that ReinBo performed better and more stably than other algorithms in our empirical experiments. Besides comparing the final performance, it is also interesting to look into the suggested machine learning pipelines by an AutoML system. The frequencies of the operators in the pipelines suggested by ReinBo are listed in Table \ref{tab:models}.

\begin{table}
	\centering
	\caption{Frequency of operators suggested by ReinBo. During empirical experiments there are 500 pipelines in total suggested by ReinBo at the end of optimization process. The frequency (Freq.) and relative frequency (Relative freq.) of each operator selected in best pipelines are shown here.}
	\footnotesize
	\scalebox{0.7}{
		\begin{tabular}{lcc|lcc|lcc}	
			\toprule
			\textbf{Preprocess} & \textbf{Freq.} & \textbf{Relative freq.} & \textbf{Feature engineering} & \textbf{Freq.} & \textbf{Relative freq.} & \textbf{Classifier} & \textbf{Freq.} & \textbf{Relative freq.} \\
			\midrule
			Scale(default)& 259& 51.8\%& FilterAnova& 210& 42.0\%& ksvm& 276& 55.2\%\\
			Scale(scale=FALSE)& 106& 21.2\%& FilterKruskal& 139& 27.8\%& ranger& 201& 40.2\%\\ 
			Scale(center=FALSE)& 67& 13.4\%& PCA& 63& 12.6\%& kknn& 12& 2.4\%\\
			NA& 36& 7.2\%& Univariate& 46& 9.2\%& xgboost& 10& 2.0\%\\
			SpatialSign& 32& 6.4\%& NA& 42& 8.4\%& naiveBayes& 1& 0.2\%\\		   
			\bottomrule
	\end{tabular}}
\label{tab:models}
\end{table}

\begin{figure}
\centering
\includegraphics[scale=.3]{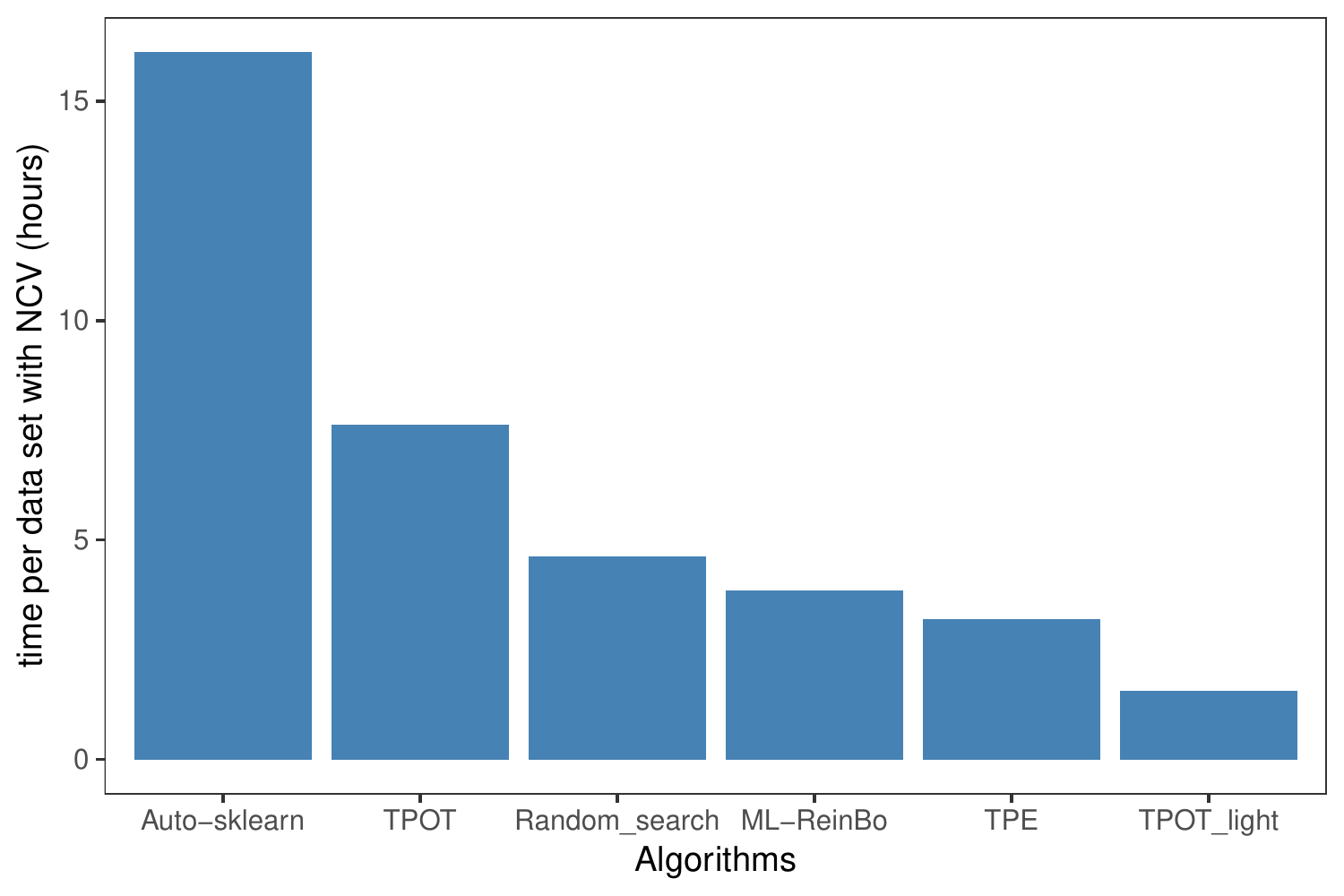} 
\caption{Barplot of the running time for each algorithm in average. The value reported here corresponds to the average running time of each algorithm per data set based on the \textit{NCV} strategy.}
\label{fig:time}
\end{figure}

\textbf{Running time}\,
Figure \ref{fig:time} shows the average running time each algorithm takes to complete one experiment, which corresponds to a complete Nested Cross Validation (\textit{NCV}) process.

From the figure, it can be seen that Auto-sklearn is the most time-consuming algorithm in our empirical experiments. Although TPOT-light is the fastest algorithm, it resulted in worse performance because it contains only fast operators. Our proposed ReinBo algorithm spent less time than Random Search and state of art AutoML systems TPOT and Auto-sklearn in average. 

\section{Summary and Future Work}
We present a new AutoML algorithm ReinBo by embedding Bayesian Optimization into Reinforcement Learning. The Reinforcement Learning takes care of pipeline composition, and Bayesian Optimization takes care of configuring the hyper-parameters associated with the composed pipeline. ReinBo is inspired by Hyperband and  previous efforts in AutoML by considering the trade-off of assigning resources to a particular configuration and exploring more configurations as a reinforcement learning problem, where the learned policy solves the trade-off automatically. Experiments show our method has a considerable improvement compared to other state of art systems and methods.
For future work, it would be interesting to include meta learning into our system, which does not only learn how to construct a pipeline and configure it for a dataset in question, but also how to generalize the learned policy to a wide range of dataset by learning jointly on the meta features.
Additionally, it would be nice to see how ReinBo performs on jointly optimizing neural architecture and continuous hyper-parameters like learning rate and momentum, as well as applications like Computer Vision \cite{jiulong2017detecting} and semantic web based Recommendation Systems \cite{kushwaha2018lesson} where pipeline might play a role. Multi-Objective Bayesian Optimization \cite{horn2017first} for the hyper-parameter tuning would be another future direction.

\section{Acknowledgement}
Janek Thomas has given us a lot of helpful suggestions, Martin Binder has helped us a lot with the mlrCPO setting, Florian Pfisterer has helped us with the Auto-sklearn setup. We thank all the above and Tieu-binh Ly for giving us a lot of suggestions for the language.
\bibliographystyle{splncs04}
\bibliography{references}

\begin{thebibliography}{10}
\providecommand{\url}[1]{\texttt{#1}}
\providecommand{\urlprefix}{URL }
\providecommand{\doi}[1]{https://doi.org/#1}

\bibitem{reticulate}
Allaire, J., Ushey, K., Tang, Y.: reticulate: Interface to 'Python' (2019),
  \url{https://CRAN.R-project.org/package=reticulate}, r package version 1.11.1

\bibitem{baker2017designing}
Baker, B., Gupta, O., Naik, N., Raskar, R.: Designing neural network
  architectures using reinforcement learning. International Conference on
  Learning Representations  (2017)

\bibitem{banzhaf1998genetic}
Banzhaf, W., Nordin, P., Keller, R.E., Francone, F.D.: Genetic programming: an
  introduction, vol.~1. Morgan Kaufmann San Francisco (1998)

\bibitem{barto2003recent}
Barto, A.G., Mahadevan, S.: Recent advances in hierarchical reinforcement
  learning. Discrete event dynamic systems  \textbf{13}(1-2),  41--77 (2003)

\bibitem{bergstra2012random}
Bergstra, J., Bengio, Y.: Random search for hyper-parameter optimization.
  Journal of Machine Learning Research  \textbf{13}(Feb),  281--305 (2012)

\bibitem{bergstra2013hyperopt}
Bergstra, J., Yamins, D., Cox, D.D.: Hyperopt: A python library for optimizing
  the hyperparameters of machine learning algorithms. In: Proceedings of the
  12th Python in science conference. pp. 13--20. Citeseer (2013)

\bibitem{bergstra2011algorithms}
Bergstra, J.S., Bardenet, R., Bengio, Y., K{\'e}gl, B.: Algorithms for
  hyper-parameter optimization. In: Advances in neural information processing
  systems. pp. 2546--2554 (2011)

\bibitem{cpo}
Binder, M.: mlrCPO: Composable Preprocessing Operators and Pipelines for
  Machine Learning (2019), \url{https://CRAN.R-project.org/package=mlrCPO}, r
  package version 0.3.4-2

\bibitem{bischl2017openml}
Bischl, B., Casalicchio, G., Feurer, M., Hutter, F., Lang, M., Mantovani, R.G.,
  van Rijn, J.N., Vanschoren, J.: Openml benchmarking suites and the openml100.
  arXiv preprint arXiv:1708.03731  (2017)

\bibitem{mlr}
Bischl, B., Lang, M., Kotthoff, L., Schiffner, J., Richter, J., Studerus, E.,
  Casalicchio, G., Jones, Z.M.: {mlr}: Machine learning in r. Journal of
  Machine Learning Research  \textbf{17}(170), ~1--5 (2016),
  \url{http://jmlr.org/papers/v17/15-066.html}

\bibitem{bischl2017mlrmbo}
Bischl, B., Richter, J., Bossek, J., Horn, D., Thomas, J., Lang, M.: mlrmbo: A
  modular framework for model-based optimization of expensive black-box
  functions. arXiv preprint arXiv:1703.03373  (2017)

\bibitem{brochu2010tutorial}
Brochu, E., Cora, V.M., De~Freitas, N.: A tutorial on bayesian optimization of
  expensive cost functions, with application to active user modeling and
  hierarchical reinforcement learning. arXiv preprint arXiv:1012.2599  (2010)

\bibitem{dietterich2000hierarchical}
Dietterich, T.G.: Hierarchical reinforcement learning with the maxq value
  function decomposition. Journal of Artificial Intelligence Research
  \textbf{13},  227--303 (2000)

\bibitem{drori2018alphad3m}
Drori, I., Krishnamurthy, Y., Rampin, R., de~Paula~Lourenco, R., Ono, J.P.,
  Cho, K., Silva, C., Freire, J.: Alphad3m: Machine learning pipeline
  synthesis. In: AutoML Workshop at ICML (2018)

\bibitem{feurer2015efficient}
Feurer, M., Klein, A., Eggensperger, K., Springenberg, J., Blum, M., Hutter,
  F.: Efficient and robust automated machine learning. In: Advances in Neural
  Information Processing Systems. pp. 2962--2970 (2015)

\bibitem{horn2017first}
Horn, D., Dagge, M., Sun, X., Bischl, B.: First investigations on noisy
  model-based multi-objective optimization. In: International Conference on
  Evolutionary Multi-Criterion Optimization. pp. 298--313. Springer (2017)

\bibitem{hutter2011sequential}
Hutter, F., Hoos, H.H., Leyton-Brown, K.: Sequential model-based optimization
  for general algorithm configuration. In: International Conference on Learning
  and Intelligent Optimization. pp. 507--523. Springer (2011)

\bibitem{jiulong2017detecting}
Jiulong, Z., Luming, G., Su, Y., Xudong, S., Xiaoshan, L.: Detecting chinese
  calligraphy style consistency by deep learning and one-class svm. In: 2017
  2nd International Conference on Image, Vision and Computing (ICIVC). pp.
  83--86. IEEE (2017)

\bibitem{kulkarni2016hierarchical}
Kulkarni, T.D., Narasimhan, K., Saeedi, A., Tenenbaum, J.: Hierarchical deep
  reinforcement learning: Integrating temporal abstraction and intrinsic
  motivation. In: Advances in neural information processing systems. pp.
  3675--3683 (2016)

\bibitem{kushwaha2018lesson}
Kushwaha, N., Sun, X., Singh, B., Vyas, O.: A lesson learned from pmf based
  approach for semantic recommender system. Journal of Intelligent Information
  Systems  \textbf{50}(3),  441--453 (2018)

\bibitem{li2017hyperband}
Li, L., Jamieson, K., DeSalvo, G., Rostamizadeh, A., Talwalkar, A.: Hyperband:
  A novel bandit-based approach to hyperparameter optimization. The Journal of
  Machine Learning Research  \textbf{18}(1),  6765--6816 (2017)

\bibitem{mnih2015human}
Mnih, V., Kavukcuoglu, K., Silver, D., Rusu, A.A., Veness, J., Bellemare, M.G.,
  Graves, A., Riedmiller, M., Fidjeland, A.K., Ostrovski, G., et~al.:
  Human-level control through deep reinforcement learning. Nature
  \textbf{518}(7540), ~529 (2015)

\bibitem{mohr2018ml}
Mohr, F., Wever, M., H{\"u}llermeier, E.: Ml-plan: Automated machine learning
  via hierarchical planning. Machine Learning  \textbf{107}(8-10),  1495--1515
  (2018)

\bibitem{olson2016tpot}
Olson, R.S., Moore, J.H.: Tpot: A tree-based pipeline optimization tool for
  automating machine learning. In: Workshop on Automatic Machine Learning. pp.
  66--74 (2016)

\bibitem{scikit-learn}
Pedregosa, F., Varoquaux, G., Gramfort, A., Michel, V., Thirion, B., Grisel,
  O., Blondel, M., Prettenhofer, P., Weiss, R., Dubourg, V., Vanderplas, J.,
  Passos, A., Cournapeau, D., Brucher, M., Perrot, M., Duchesnay, E.:
  Scikit-learn: Machine learning in {P}ython. Journal of Machine Learning
  Research  \textbf{12},  2825--2830 (2011)

\bibitem{de2017recipe}
de~S{\'a}, A.G., Pinto, W.J.G., Oliveira, L.O.V., Pappa, G.L.: Recipe: a
  grammar-based framework for automatically evolving classification pipelines.
  In: European Conference on Genetic Programming. pp. 246--261. Springer (2017)

\bibitem{silver2017mastering}
Silver, D., Schrittwieser, J., Simonyan, K., Antonoglou, I., Huang, A., Guez,
  A., Hubert, T., Baker, L., Lai, M., Bolton, A., et~al.: Mastering the game of
  go without human knowledge. Nature  \textbf{550}(7676), ~354 (2017)

\bibitem{sun}
Sun, X., Bommert, A., Pfisterer, F., Rahnenführer, J., Lang, M., Bischl, B.:
  High dimensional restrictive federated model selection with multi-objective
  bayesian optimization over shifted distributions (2019)

\bibitem{DBLP:journals/corr/abs-1208-3719}
Thornton, C., Hutter, F., Hoos, H.H., Leyton{-}Brown, K.: Auto-weka: Automated
  selection and hyper-parameter optimization of classification algorithms. CoRR
   \textbf{abs/1208.3719} (2012), \url{http://arxiv.org/abs/1208.3719}

\bibitem{thornton2012auto}
Thornton, C., Leyton-Brown, K.: Auto-weka: Automated selection and
  hyper-parameter optimization of classification algorithms  (2012)

\bibitem{watkins1992q}
Watkins, C.J., Dayan, P.: Q-learning. Machine learning  \textbf{8}(3-4),
  279--292 (1992)

\bibitem{yang2019program}
Yang, F., Gustafson, S., Elkholy, A., Lyu, D., Liu, B.: Program search for
  machine learning pipelines leveraging symbolic planning and reinforcement
  learning. In: Genetic Programming Theory and Practice XVI, pp. 209--231.
  Springer (2019)

\bibitem{yang2018peorl}
Yang, F., Lyu, D., Liu, B., Gustafson, S.: Peorl: Integrating symbolic planning
  and hierarchical reinforcement learning for robust decision-making. arXiv
  preprint arXiv:1804.07779  (2018)

\bibitem{zoph2016neural}
Zoph, B., Le, Q.V.: Neural architecture search with reinforcement learning.
  arXiv preprint arXiv:1611.01578  (2016)

\end{thebibliography}
\end{document}